\documentclass[sigconf]{acmart}
\AtBeginDocument{%
  \providecommand\BibTeX{{%
    \normalfont B\kern-0.5em{\scshape i\kern-0.25em b}\kern-0.8em\TeX}}}

\copyrightyear{2022}
\acmYear{2022}
\setcopyright{acmcopyright}
\acmConference[MuSe' 22]{Proceedings of the 3rd International Multimodal Sentiment Analysis Workshop and Challenge}{October 10, 2022}{Lisboa, Portugal}
\acmBooktitle{Proceedings of the 3rd International Multimodal Sentiment Analysis Workshop and Challenge (MuSe' 22), October 10, 2022, Lisboa, Portugal}
\acmPrice{15.00}
\acmDOI{10.1145/3551876.3554802}
\acmISBN{978-1-4503-9484-0/22/10}

\begin{document}

\title{Hybrid Multimodal Fusion for Humor Detection}

\author{Haojie Xu}
\affiliation{%
 \institution{AHU-IAI AI Joint Laboratory, \\Anhui University}
 \institution{Institute of 
Artificial Intelligence, \\Hefei Comprehensive National Science Center}
 \city{Hefei}
 \country{China}}
\email{e20201139@stu.ahu.edu.cn}

\author{Weifeng Liu}
\affiliation{%
 \institution{AHU-IAI AI Joint Laboratory, \\Anhui University}
 \institution{Institute of 
Artificial Intelligence, \\Hefei Comprehensive National Science Center}
 \city{Hefei}
 \country{China}}
\email{wfeng_ch@163.com}

\author{Jingwei Liu}
\affiliation{%
 \institution{AHU-IAI AI Joint Laboratory, \\Anhui University}
 \institution{Institute of 
Artificial Intelligence, \\Hefei Comprehensive National Science Center}
 \city{Hefei}
 \country{China}}
\email{ljw578616559@163.com}

\author{Mingzheng Li}
\affiliation{%
 \institution{University of Science and Technology of China}
 \institution{Institute of 
Artificial Intelligence, \\Hefei Comprehensive National Science Center}
 \city{Hefei}
 \country{China}}
\email{mingzhengli@mail.ustc.edu.cn}

\author{Yu Feng}
\affiliation{%
 \institution{School of Biomedical Engineering, \\Anhui Medical University}
 \institution{Institute of 
Artificial Intelligence, \\Hefei Comprehensive National Science Center}
 \city{Hefei}
 \country{China}}
\email{fengyu1919@126.com}

\author{Yasi Peng}
\affiliation{%
 \institution{School of Biomedical Engineering, \\Anhui Medical University}
 \institution{Institute of 
Artificial Intelligence, \\Hefei Comprehensive National Science Center}
 \city{Hefei}
 \country{China}}
\email{1558778695@qq.com}

\author{Yunwei Shi}
\affiliation{%
 \institution{AHU-IAI AI Joint Laboratory, \\Anhui University}
 \institution{Institute of 
Artificial Intelligence, \\Hefei Comprehensive National Science Center}
 \city{Hefei}
 \country{China}}
\email{1826719913@qq.com}

\author{Xiao Sun}
\authornote{Corresponding author.}
\affiliation{%
 \institution{Hefei University of Technology}
 \institution{ZhongJuYuan Intelligent Technology Co.,\\Ltd}
 \city{Hefei}
 \country{China}}
\email{sunx@iai.ustc.edu.cn}

\author{Meng Wang}
\affiliation{%
 \institution{Hefei University of Technology}
 \institution{Institute of 
Artificial Intelligence, \\Hefei Comprehensive National Science Center}
 \city{Hefei}
 \country{China}}
\email{eric.mengwang@gmail.com}


\begin{abstract}

In this paper, we present our solution to the MuSe-Humor sub-challenge of the Multimodal Emotional Challenge (MuSe) 2022. The goal of the MuSe-Humor sub-challenge is to detect humor and calculate AUC from audiovisual recordings of German football Bundesliga press conferences. It is annotated for humor displayed by the coaches. For this sub-challenge, we first build a discriminant model using the transformer module and BiLSTM module, and then propose a hybrid fusion strategy to use the prediction results of each modality to improve the performance of the model. Our experiments demonstrate the effectiveness of our proposed model and hybrid fusion strategy on multimodal fusion, and the AUC of our proposed model on the test set is 0.8972.
  
\end{abstract}




\maketitle

\section{Introduction}
Humor is the absurd, unexpected, yet subtle or evocative characteristic of something. It is also a way to help people relieve stress. Not only does it enhance feelings, but it also helps increase overall well-being. As a popular research field in natural language processing (NLP), humor detection has received more attention from scholars \cite{DBLP:conf/wilf/BuscaldiR07,DBLP:conf/acl/LiuZS18,yang2015humor,DBLP:journals/cluster/Zhu19b}. 

The inconsistency theory of humor argues that humor arises from two or more incongruent but related situations. However, due to differences in thought, culture and cognition, people\textquotesingle s understanding of humor is not the same. This means that humor detection requires a lot of prior knowledge and background information, which brings great challenges for the machine to understand humor. Fortunately, humor often comes from people's interactions. In addition to the text containing the context, one can observe the expression of the speaker in the process of speaking, and we can also obtain the rhythmic clues in his voice. All of this information contributes to the detection of the humorous element. In other words, utilizing the multimodal characteristics of text, acoustic and visual can greatly help the humor detection task \cite{DBLP:conf/emnlp/HasanRZZTMH19,DBLP:conf/icmla/AbdullahHS18}.

In general, multimodal feature fusion includes early-fusion and late-fusion. For early-fusion, the characteristics of all modalities are fused at first, then they are sent to the model for training. As there are many differences between different modalities, mixing them directly may lead to the underutilization of information. With regard to the late-fusion method, researchers first utilize certain individual models to extract features of each modality. Then, the features being extracted are combined for subsequent tasks. However, the training process may result in the loss of original information. 
Therefore, we propose a hybrid multimodal fusion model for humor detection, called HMF-MD, which absorbs the advantages of the two fusion methods. First, several discriminant modules are applied to every single modality. Then we combine these features with the original data and send them into another discriminant module for humor detection. This can not only extract the key information inside each modality but also prevent the loss of information effectively. Experiments show that our model achieves good results on the MuSe-Humor sub-challenge task.

Specifically, the main contributions of our work are as follows:
\begin{itemize}
    \item We propose a discriminative module, which can extract contextual and key information from a single modality.
    \item We propose a new hybrid fusion strategy that can improve model performance.
    \item Experiments demonstrate the effectiveness of our proposed model and hybrid fusion strategy on multimodal fusion, and the AUC of our proposed model on the test set is 0.8972.
\end{itemize}

\section{RELATED WORKS}
\subsection{Humor Detection}
Humor detection has been one of the active areas in the field of affective computing. Humor recognition is to identify whether a sentence or an utterance is humorous or not. There are many ways to be used to identify humor. \cite{yang2015humor} uses non-neural models to recognize humor. Recently, there are some studies using recurrent neural networks (RNNs) and convolutional neural networks (CNNs) to detect humor\cite{de2015humor}. The pre-trained language model has achieved great success in many areas, there are also a lot of works using transformer-based architecture and attention mechanisms to detect humor\cite{annamoradnejad2020colbert,mao2019bert}.

 Multimodal humor detection is a new area of research in NLP. Multimodal studies from textual, visual and acoustic are the recent research trends. Many works present new Multimodal neural architectures \cite{poria2017multi,hazarika2018conversational,related69}, and multimodal fusion approaches
\cite{barezi2018modality,liang2018multimodal}.
\subsection{Multimodal Fusion}
In a multimodal setting where multiple modalities convey information from different channels, cross-modal fusion is crucial in exploring intra-modality and inter-modality dynamics to mine complementary information. In recent years, multimodal fusion has seen rapid development mainly thanks to the multimodal machine learning community. Earlier multimodal fusion methods fall into two broad categories, i.e. feature-level fusion and decision-level fusion, otherwise known as early and late fusion respectively. Feature-level fusion methods mostly fuse features through concatenation of unimodal representations \cite{related13,related14,related15}, whereas decision-level fusion methods firstly make a tentative inference for each modality and further fuse them using a voting mechanism \cite{related18,related63, related64,related65,related66}. However, these two types of fusion methods cannot effectively explore the inter-modality dynamics. Recently proposed fusion methods can be categorized into several types as follows, i.e., multi-view learning methods \cite{related67,related68}, word-level fusion methods \cite{related69,related70,related71}, tensor fusion \cite{related19,related20},  and hybrid fusion\cite{hm11,hm12}. These fusion techniques are effective in learning inter-modality dynamics compared to feature-level and decision-level fusion and show marked performance gains.

\section{MULTIMODAL FEATURES}
\subsection{Acoustic Features}
All audio files are first normalised to -3 decibels and then converted from stereo to mono, at 16 kHz, 16 bit. Afterwards, we make use of the two well-established machine learning toolkits openSMILE\cite{eyben2010opensmile} and DeepSpectrum\cite{amiriparian2017snore} for expert-designed and deep feature extraction from the audio recordings.\par

\textbf{eGeMAPS Feature}: We use the extended Geneva Minimalistic Acoustic Parameter Set (eGeMAPS) feature provided by the organizers of MuSe 2022, which contains 23 acoustic low-level descriptors (LLDs)\cite{eyben2015geneva}. The freely and publicly available openSMILE toolkit can be used to extract the eGeMAPS feature. Several statistical functions of the openSMILE toolkit can be directly applied to extract segment-level features with an 88-dimensional vector.\par
\textbf{DeepSpectrum Feature}: The principle of DeepSpectrum is to utilise pre-trained image Convolutional Neural Networks (CNNs) for the extraction of deep features from visual representations (e. g., Mel-spectrograms) of audio signals.

\subsection{Visual Features}
To extract specific image descriptors related to facial expressions, we make use of two CNN architectures: Multi-task Cascaded Convolutional Networks (MTCNN)\cite{zhang2016joint} and VGGface 2\cite{cao2018vggface2}. We also provide a set of Facial Action Units (FAUs) obtained from faces of individuals in the datasets. \par

\textbf{MTCNN}: The MTCNN model, pre-trained on the datasets WIDER FACE\cite{yang2016wider} and CelebA\cite{liu2015deep}, is used to detect faces in the videos. The extracted faces then serve as inputs of the feature extractors VGGface 2.\par
\textbf{VGGface 2}: The purpose of VGGface 2 is to compute general facial features for the previously extracted faces. VGGface 2 is a dataset for the task of face recognition. It contains 3.3 million faces of about 9,000 different persons. We use a ResNet50\cite{he2016deep} trained on VGGface 2 and detach its classification layer, resulting in a 512-dimensional feature vector output referred to as VGGface 2.\par
\textbf{FAUs}: Facial Action Units (FAUs) denote the presence of facial muscle movements that are commonly used for describing and classifying expressions. It can be extracted by OpenFace toolkit \cite{baltruvsaitis2016openface}. We only use the FAU intensity features provided by the organizers.

\subsection{Textual Features}
\textbf{Bert}: As the organizer did\cite{baseline,baseline1}, We employ a German version of the BERT (Bidirectional Encoder Representations from Transformers (BERT)\cite{devlin2018bert}) model. No further fine-tuning is applied. For Passau Spontaneous Football Coach Humor (Passau-SFCH), the dataset used for the humor detection sub-challenge, we extract the BERT token embeddings. Additionally, we obtain 768-dimensional sentence embeddings for all texts in Passau-SFCH by using the encodings of BERT ’s token. In all cases, we average the embeddings provided by the last 4 layers of the BERT model, following\cite{sun2020multi}.

\section{MUSE-Humor METHOD}
\begin{figure}[h]
  \centering
  \includegraphics[width=\linewidth]{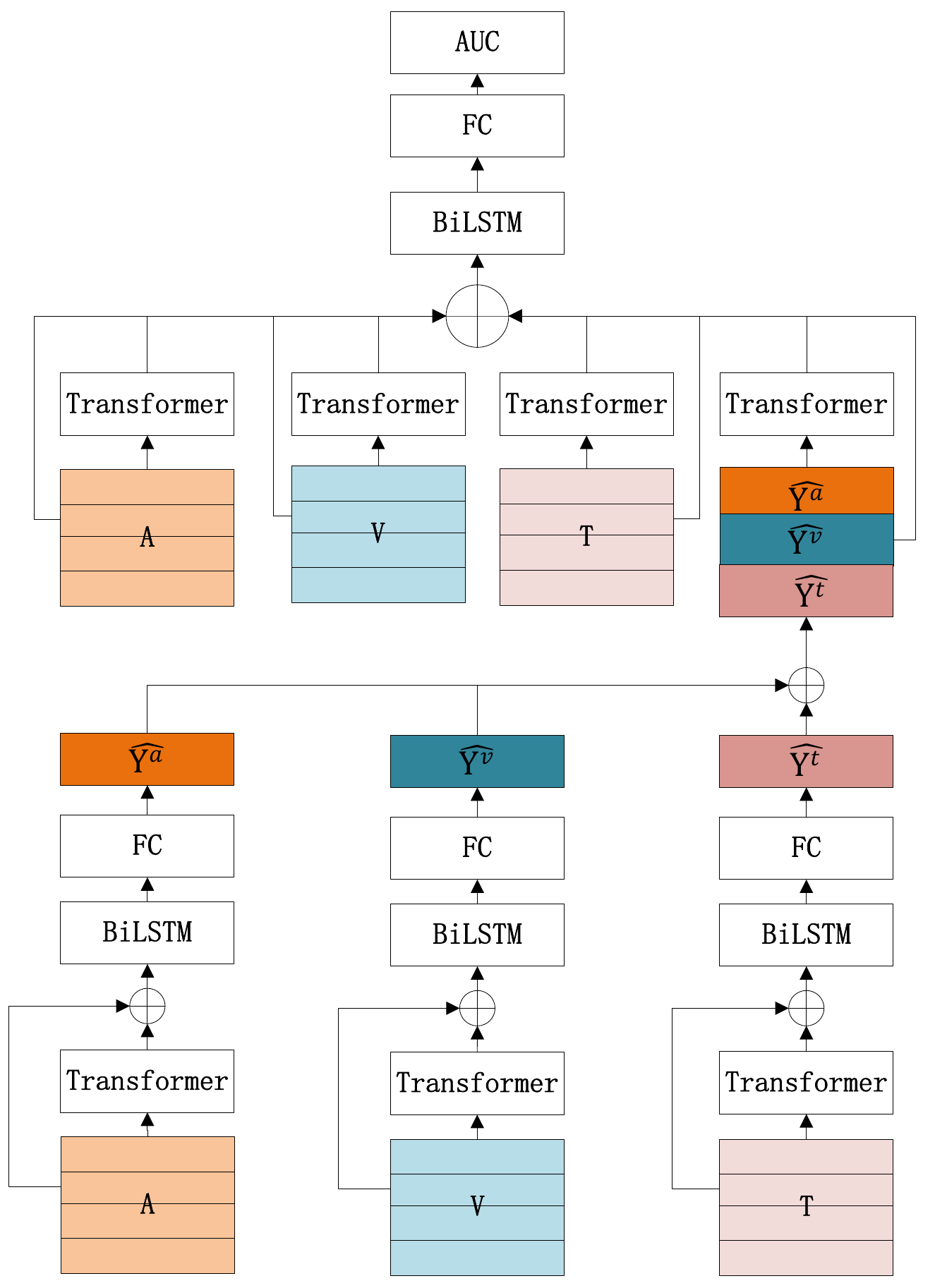}
  \caption{Overview of the architecture used in the MuSe-Humor sub-challenge.}
    \label{fig:model}
\end{figure}
In this section, we will introduce the main components of our approach, and its overall framework is shown in Figure \ref{fig:model}. We propose a hybrid multimodal fusion model for humor detection, called HMF-MD. The training process of HMF-MD consists of two stages: (1) Unimodal Discrimination Stage, which obtains the optimal discriminant distribution of the input sequence corresponding to each modality through our discriminant module. (2) Hybrid Multimodal Fusion Stage, which utilizes the Hybrid Multimodal Fusion strategy to fuse the initial input of various modalities and the discriminant output of every single modality in the previous stage, and finally performs humor detection through our discriminant module.

\subsection{Unimodal Discrimination Stage}\label{sec:4.1}
For this humor detection sub-challenge, the data is not monomodal, so interactions of multimodal information are inevitably considered. However, for data of a particular modality, there are internal correlations specific to that modality, so utilizing single modality data for the effective fusion of multimodal information is essential.

To capture this correlation within a certain modality, we propose the discriminant module shown in the bottom half of Figure \ref{fig:model}. Given an input sequence $X^m=\{x_1^m,x_2^m,\dots,x_L^m\}$ for a specific modality $m\in\{A,V,T\}$, where $L$ is the sequence length, we first feed it into a Transformer layer\cite{attention} to capture the interaction between each element in the sequence and other elements and retain more information about the element itself through a residual connection.
\begin{equation}\label{eq1}
    H^m = \mathcal{T}(X^m)\oplus X^m
\end{equation}
where $H^m=\{h^m_1,h^m_2,\dots,h^m_L\}$ is the hidden representation sequence output by Eq.\ref{eq1}, $\mathcal{T}(\cdot)$ denotes the calculation process in the Transformer layer\cite{attention}, and $m\in\{A,V,T\}$ represents a particular modality.

Since the data of the humor detection sub-challenge has an obvious contextual relationship, we then send the obtained hidden representation sequence into a  bidirectional LSTM \cite{lstm} to capture this contextual information and take the hidden output of the last element in the sequence as the hidden representation of this sequence.
\begin{equation}
    \tilde{H}^m=\overrightarrow{\mathcal{B}^m}(H^m)[-1]||\overleftarrow{\mathcal{B}^m}(H^m)[0]
\end{equation}
where $\tilde{H}^m$ is output of the BiLSTM layer, $\overrightarrow{\mathcal{B}}(\cdot)$ and $\overleftarrow{\mathcal{B}}(\cdot)$ denote the forward and backward calculation process of LSTM, $[-1]$ and $[0]$ represent the last and first and elements of the corresponding output sequence, and $||$ denotes the concatenate operation.

Finally, we send it to the last component of the discrimination module, a fully connected layer, and its output is the representation that captures the internal correlation of the corresponding modality, i.e. the discriminant output of each modality.
\begin{equation}
    \hat{Y}^m=\sigma(W^m\tilde{H}^m+b^m)
\end{equation}
where $W^m$ and $b^m$ are the modality-specific weight matrix and the bias term, respectively, and $\sigma$ is the Sigmoid activation function. 

For the input sequence of each modality, we carry out a complete training process to obtain the optimal discriminant output.

\subsection{Hybrid Multimodal Fusion Stage}
In order to integrate the information of various modality more effectively and improve the advantages brought by information complementarity between modalities, we propose the hybrid fusion strategy at this stage.

Firstly, we concatenate the discriminant outputs of each modality in the first stage and take them as the input of the second stage. 
\begin{equation}
    X^{\hat{Y}}=||_{m\in\{A,V,T\}}\hat{Y}^m
\end{equation}

However, as described in Section \ref{sec:4.1}, these discriminant outputs are modality-specific internal correlations, and simply using them for the final detection task cannot take full advantage of the complementary effects between the various modalities. Therefore, in addition to the output of the first stage, the original input of each modality is also taken as the input feature of the second stage, so that the model can effectively capture the complementary information of each modality.

The main structure of the second stage is still the discrimination module introduced in Section \ref{sec:4.1}. The difference is that in the second stage, each type of input first passes through its own transformer layer first, and then the combined output is sent to the shared subsequent layers for humor detection.
\begin{equation}
    Z = ||_{m\in\{A,V,T,{\hat{Y}}\}}\mathcal{T}^m(X^m)\oplus X^m
\end{equation}
\begin{equation}
    \tilde{Z}=\overrightarrow{\mathcal{B}}(Z)[-1]||\overleftarrow{\mathcal{B}}(Z)[0]
\end{equation}
\begin{equation}
    \hat{Z}=\sigma(W\tilde{Z}+b)
\end{equation}

\subsection{Wrapped BCELoss}
In both stage, we train our model using the Wrapped BCELoss, which can be formatted as follows:
\begin{equation}
    \ell = \dfrac{1}{N}\sum_{n=1}^N l_n
\end{equation}
\begin{equation}
    l_n = - w_n \left[y_n \cdot \log x_n + (1 - y_n) \cdot \log (1 - x_n) \right]
\end{equation}
where $N$ is the total number of samples, $l_n$ is the corresponding loss of the $n\text{-th}$ sample, and $x_n$ and $y_n$ are the ground-truth label and the predicted label for the $n\text{-th}$ sample.

\section{EXPERIMENTS}

\begin{table}
	\caption{Statistics information of the Passau-SFCH dataset.}
	\label{tab:dataset}
	\begin{tabular}{c|ccc}
		\toprule
		Partition& Coaches &Labels& Duration\\
		\midrule
		Train & 4 & 14025& 3 :52 :44\\
		Development & 3 & 11320& 3 :08 :12\\
		Test & 3 & 14143& 3 :55 :41\\
		\midrule
		Sum & 10 & 39488&10 :56 :37\\
		\bottomrule
	\end{tabular}
\end{table}

\begin{table}
	\caption{ The AUC performance on the development set of the MuSe-Humor sub-challenge.
		\lq{M}{\rq} denotes the used modality.
	 \lq{A}{\rq}, \lq{V}{\rq} and \lq{T}{\rq} represent the audio, video, and text modality. \lq{BiLSTM}{\rq} means the official baseline.}
	\label{tab:1}
	\begin{tabular}{cc|cc}
		
		\toprule
		Feature&M&Model&AUC\\
		\midrule
		eGeMAPS & A &BiLSTM&0.6861 \\
		eGeMAPS & A &Ours&0.6912 \\
		DeepSpectrum & A &BiLSTM& 0.7149\\
        DeepSpectrum & A &Ours& 0.7187\\
		\midrule
		FAU  & V &BiLSTM& 0.9071\\
		FAU & V &Ours& 0.9050\\
		VGGface 2 & V &BiLSTM& 0.9253\\
		VGGface 2& V &Ours& 0.9331\\
		\midrule
		BERT & T &BiLSTM& 0.8270\\
		BERT & T&Ours& 0.8261\\
		\bottomrule
	\end{tabular}
\end{table}

\begin{table*}
	\caption{AUC performance of different modalities using hybrid fusion strategy on the development set in the MuSe-Humor sub-challenge. $\hat{Y}$ represents the discriminative output of the corresponding mode. $\hat{Y^{m}}$ represents the discriminative output obtained by using BiLSTM for the corresponding modality. $\hat{Y^{M}}$ represents the discriminative output obtained by using our proposed model for the corresponding modality.}
	\label{tab:3}
	\begin{tabular}{c|c|c|c}
		\toprule
		Feature&M&Model&AUC\\
		\midrule
		DeepSpectrum + VGGface& A+V&BiLSTM  & 0.8252\\
		DeepSpectrum + VGGface& A+V&Ours  & 0.9306\\
		DeepSpectrum + VGGface & A+V+$\hat{Y^{A,V}}$&Ours  & 0.9415\\
		\midrule
		DeepSpectrum + BERT & A+T&BiLSTM& 0.8901\\
		DeepSpectrum + BERT & A+T&Ours& 0.8303\\
		DeepSpectrum + BERT & A+T+$\hat{Y^{A,T}}$&Ours& 0.8508\\
		\midrule
		VGGface 2 + BERT & V+T &BiLSTM& 0.8908\\
		VGGface 2 + BERT & V+T &Ours& 0.9461\\
		VGGface 2 + BERT & V+T +$\hat{Y^{V,T}}$&Ours& 0.9483\\
		\midrule
		DeepSpectrum + VGGface 2 +BERT& A+V+T&BiLSTM & 0.9033\\
		DeepSpectrum + VGGface 2 +BERT& A+V+T&Ours & 0.9534\\
		DeepSpectrum + VGGface 2 +BERT& A+V+T+$\hat{Y^{A,V,T}}$ &Ours& 0.9552\\
		DeepSpectrum + VGGface 2 +BERT& A+V+T+$\hat{Y^{A,V,t}}$ &Ours& 0.9570\\
		DeepSpectrum + VGGface 2 +BERT& A+V+T+$\hat{Y^{a,V,t}}$ &Ours& 0.9573\\
		\bottomrule
	\end{tabular}
\end{table*}

\subsection{Dataset}
In MuSe 2022, three datasets are provided for three different sub-challenges. This paper mainly focuses on the study of humor detection sub-challenge. The dataset for the humor detection sub-challenge is the novel Passau Spontaneous Football Coach Humor (Passau-SFCH) database. It comprises audiovisual recordings of German football Bundesliga press conferences. It is annotated for humor displayed by the coaches. For the challenge, binary labeling (presence or absence of humor) is provided. Each label in Passau-SFCH dataset is predicted based on all feature vectors belonging to the corresponding 2 s window.  The statistics information is shown in Table \ref{tab:dataset}.

\subsection{Experimental Setup}
We implement all of our models with the PyTorch toolkit in the MuSe-Humor sub-challenges. 
For the MuSe-Humor sub-challenge, the proposed model consists of the transformer layer, a bidirectional LSTM layer, and a fully connected layer. The number of layers used by all transformer layers is 1, and the number of heads used by all Multi-head attention is 1. The number of hidden sizes in the LSTM layer is 32, and the number of the bidirectional LSTM layer is set to 2. The Adam optimizer with a learning rate of 0.001 is applied, and the batch size is set to 32. Same as the baseline, we train the model for a maximum of 100 epochs and stop training if the validation AUC does not increase for 3 consecutive epochs. For the hybrid fusion model, we only use 0.4 for the dropout of the linear layer and 0.2 for the rest of the modules.

\subsection{Unimodal Results}

We first evaluate the performance of each modality we used in the MuSe-Humor sub-challenge. To verify the effectiveness of the proposed model, several experiments are conducted. The experiment results are given in Table \ref{tab:1}. From the table \ref{tab:1}, we can conclude that: 1) The \lq{Ours}{\rq} model achieves the best performance or performance close to \lq{BiLSTM}{\rq} when using audio and video modalities for classification. 2) \lq{BiLSTM}{\rq} model outperforms \lq{Ours}{\rq} models when using text for classification. We believe that the reason is that the feature continuity has deteriorated during feature extraction and processing, and better sequence information can be obtained by directly feeding it into BiLSTM. Nevertheless, the performance difference between \lq{Ours}{\rq} and \lq{BiLSTM} is very small. 3)On the whole, \lq{Ours}{\rq} can achieve good performance in the independent use of the three modalities.

\subsection{Multimodal Results}


When performing multi-modal feature fusion, we select the same modal features as the official baseline for fusion. Compared with the official baseline, we also get better results on the single modality for text features (BERT), audio features (DeepSpectrum) and visual features (VGGface 2). Table \ref{tab:3} shows the AUC performance of different modalities with and without the hybrid fusion strategy on the development set of the MuSe-Humor sub-challenge.
To better verify the effectiveness of our proposed hybrid fusion strategy, we list the discriminative output $\hat{Y}$ as a new modality in Table \ref{tab:3}. For example, \lq{A+V+T}{\rq} means that the features of these three modalities are directly sent to different Transformer layers without using a hybrid fusion strategy, and then the connected features are sent to the BiLSTM layer. \lq{A+V+T+$\hat{Y^{A,V,T}}$}{\rq} means using a hybrid fusion strategy, that is, the discriminative output $\hat{Y^{A,V,T}}$ is obtained first, and then the three modalities and the discriminative outputs obtained from the three modalities are sent to different Transformer layers. 

When we perform feature fusion of the three modalities, we also utilize BiLSTM to obtain discriminative outputs for textual modalities and acoustic modalities.
 In Table \ref{tab:3}, \lq{A+V+T+$\hat{Y^{A,V,t}}$}{\rq} means that BiLSTM obtains the discriminative output of text modality, and our proposed model also obtains the discriminative output of acoustic and visual modality. 
 \lq{A+V+T+$\hat{Y^{a,V,t}}$}{\rq} means that BiLSTM obtains the discriminative output of textual and acoustic modality, and our proposed model also obtains the discriminative output of visual modality.

From Table \ref{tab:3}, we can find that 1) multi-modal features can achieve better performance than single-modal features. 2) In the results of our model, the results with the hybrid fusion strategy are all better than those without the hybrid fusion strategy. 3) When multi-modal feature fusion is performed, better performance can be obtained by the fusion of video and text features or audio and video features, which is consistent with the phenomenon that the video feature results are optimal in a single modality. 4) When the audio and text features are fused, we find that the model results are lower than the official baseline BiLSTM, which we suspect is related to feature extraction and processing. Because each label is predicted based on all feature vectors belonging to the corresponding 2s window, the text features and audio features of a whole sentence may be truncated, and the direct use of BiLSTM can better maintain the temporal continuity of features. 5) Using BiLSTM to obtain the discriminative output of text modality and audio modality can improve the model performance to a certain extent.

\subsection{Submission Results}
\begin{table}
	\caption{The best submission results of our proposed method in the MuSe-Humor sub-challenges.}
	\label{tab:4}
	\begin{tabular}{c|c|c|c}
		\toprule
		Model&M&Development&Test\\
		\midrule
		Baseline& V& 0.9253& 0.8480 \\
		Baseline& A+V+T& 0.9033 & 0.7973 \\
		Ours& A+V+T& 0.9534 & 0.8559 \\
		Ours&A+V+T+$\hat{Y^{A,V,T}}$& 0.9552 & 0.8823 \\
		Ours&A+V+T+$\hat{Y^{A,V,t}}$& 0.9570& 0.8882 \\
		Ours&A+V+T+$\hat{Y^{a,V,t}}$& 0.9573& 0.8945 \\
		Ours&VOTE& -& 0.8972 \\
		\bottomrule
	\end{tabular}
\end{table}
Table \ref{tab:4} shows the best submission results of the proposed method in the MuSe-Humor sub-challenge. The official baseline achieves the best AUC results on the test set on video modality features, and the optimal AUC of our proposed model is 0.0343 higher than the official baseline. Compared with the official baseline results using three modalities, our proposed model outperforms the baseline by 0.085 in AUC.  Further, we also use \lq{A+V+T+$\hat{Y^{A,V,t}}$}{\rq}, \lq{A+V+T+$\hat{Y ^{a,V,t}}$}{\rq} and \lq{A+V+T+$\hat{Y^{A,V,T}}$}{\rq}'s prediction results to vote, and obtain the average of the predicted results. After submitting the average we reached 0.8972 on AUC.
 In addition, from the table \ref{tab:4}, we also found that adding a hybrid fusion strategy can also improve the recognition ability of the model, so that the model can achieve better results on the test set.
 When we use the hybrid fusion strategy, the model has a small performance improvement on the validation set, but the test set performance is greatly improved.
Overall, the AUC results achieve a decent performance improvement over the official baseline.
However, the AUC of the test set is still relatively different from the AUC of the validation set. We conjecture that the distribution of test and validation sets may be different and that different coaches have different habits of expressing humor, so the model may have an overfitting problem.
\section{CONCLUSIONS}
In this paper, we present our solutions for the MuSe-Humor sub-challenge of Multi-modal Sentiment Challenge (MuSe) 2022. 
For the MuSe-Humor sub-challenge, fusing the textual features (BERT) with the audio features(DeepSpectrum) and the visual features (VGGface 2) proved useful for calculating AUC. Also, our proposed hybrid fusion strategy can further improve the model performance.

The proposed method shows promising prospects for future improvements. First, more features of different modalities can be used in this sub-challenge. Then, the next step can be to explore the number of Transformer layers used by different modalities and other methods to increase the superiority and generalizability of the model. Finally, ways to maintain the temporal continuity of features during extraction and processing can be explored.
\bibliographystyle{ACM-Reference-Format}
\bibliography{sample-base}


\begin{thebibliography}{45}


\ifx \showCODEN    \undefined \def \showCODEN     #1{\unskip}     \fi
\ifx \showDOI      \undefined \def \showDOI       #1{#1}\fi
\ifx \showISBNx    \undefined \def \showISBNx     #1{\unskip}     \fi
\ifx \showISBNxiii \undefined \def \showISBNxiii  #1{\unskip}     \fi
\ifx \showISSN     \undefined \def \showISSN      #1{\unskip}     \fi
\ifx \showLCCN     \undefined \def \showLCCN      #1{\unskip}     \fi
\ifx \shownote     \undefined \def \shownote      #1{#1}          \fi
\ifx \showarticletitle \undefined \def \showarticletitle #1{#1}   \fi
\ifx \showURL      \undefined \def \showURL       {\relax}        \fi
\providecommand\bibfield[2]{#2}
\providecommand\bibinfo[2]{#2}
\providecommand\natexlab[1]{#1}
\providecommand\showeprint[2][]{arXiv:#2}

\bibitem[Abdullah et~al\mbox{.}(2018)]%
        {DBLP:conf/icmla/AbdullahHS18}
\bibfield{author}{\bibinfo{person}{Malak Abdullah}, \bibinfo{person}{Mirsad
  Hadzikadic}, {and} \bibinfo{person}{Samira Shaikh}.}
  \bibinfo{year}{2018}\natexlab{}.
\newblock \showarticletitle{{SEDAT:} Sentiment and Emotion Detection in Arabic
  Text Using {CNN-LSTM} Deep Learning}. In \bibinfo{booktitle}{\emph{17th
  {IEEE} International Conference on Machine Learning and Applications, {ICMLA}
  2018, Orlando, FL, USA, December 17-20, 2018}},
  \bibfield{editor}{\bibinfo{person}{M.~Arif Wani}, \bibinfo{person}{Mehmed~M.
  Kantardzic}, \bibinfo{person}{Moamar~Sayed Mouchaweh},
  \bibinfo{person}{Jo{\~{a}}o Gama}, {and} \bibinfo{person}{Edwin Lughofer}}
  (Eds.). \bibinfo{publisher}{{IEEE}}, \bibinfo{pages}{835--840}.
\newblock
\urldef\tempurl%
\url{https://doi.org/10.1109/ICMLA.2018.00134}
\showDOI{\tempurl}


\bibitem[Amiriparian et~al\mbox{.}(2022)]%
        {baseline1}
\bibfield{author}{\bibinfo{person}{Shahin Amiriparian}, \bibinfo{person}{Lukas
  Christ}, \bibinfo{person}{Andreas K\"onig}, \bibinfo{person}{Eva-Maria
  Meßner}, \bibinfo{person}{Alan Cowen}, \bibinfo{person}{Erik Cambria}, {and}
  \bibinfo{person}{Bj\"orn~W. Schuller}.} \bibinfo{year}{2022}\natexlab{}.
\newblock \showarticletitle{MuSe 2022 Challenge: Multimodal Humour, Emotional
  Reactions, and Stress}. In \bibinfo{booktitle}{\emph{Proceedings of the 30th
  ACM International Conference on Multimedia (MM'22), October 10-14, 2022,
  Lisbon, Portugal.}} \bibinfo{publisher}{Association for Computing Machinery},
  \bibinfo{address}{Lisbon, Portugal}.
\newblock
\newblock
\shownote{3 pages, to appear}.


\bibitem[Amiriparian et~al\mbox{.}(2017)]%
        {amiriparian2017snore}
\bibfield{author}{\bibinfo{person}{Shahin Amiriparian},
  \bibinfo{person}{Maurice Gerczuk}, \bibinfo{person}{Sandra Ottl},
  \bibinfo{person}{Nicholas Cummins}, \bibinfo{person}{Michael Freitag},
  \bibinfo{person}{Sergey Pugachevskiy}, \bibinfo{person}{Alice Baird}, {and}
  \bibinfo{person}{Bj{\"o}rn Schuller}.} \bibinfo{year}{2017}\natexlab{}.
\newblock \showarticletitle{Snore sound classification using image-based deep
  spectrum features}.
\newblock  (\bibinfo{year}{2017}).
\newblock


\bibitem[Annamoradnejad and Zoghi(2020)]%
        {annamoradnejad2020colbert}
\bibfield{author}{\bibinfo{person}{Issa Annamoradnejad} {and}
  \bibinfo{person}{Gohar Zoghi}.} \bibinfo{year}{2020}\natexlab{}.
\newblock \showarticletitle{Colbert: Using bert sentence embedding for humor
  detection}.
\newblock \bibinfo{journal}{\emph{arXiv preprint arXiv:2004.12765}}
  (\bibinfo{year}{2020}).
\newblock


\bibitem[Baltru{\v{s}}aitis et~al\mbox{.}(2016)]%
        {baltruvsaitis2016openface}
\bibfield{author}{\bibinfo{person}{Tadas Baltru{\v{s}}aitis},
  \bibinfo{person}{Peter Robinson}, {and} \bibinfo{person}{Louis-Philippe
  Morency}.} \bibinfo{year}{2016}\natexlab{}.
\newblock \showarticletitle{Openface: an open source facial behavior analysis
  toolkit}. In \bibinfo{booktitle}{\emph{2016 IEEE Winter Conference on
  Applications of Computer Vision (WACV)}}. IEEE, \bibinfo{pages}{1--10}.
\newblock


\bibitem[Barezi and Fung(2018)]%
        {barezi2018modality}
\bibfield{author}{\bibinfo{person}{Elham~J Barezi} {and}
  \bibinfo{person}{Pascale Fung}.} \bibinfo{year}{2018}\natexlab{}.
\newblock \showarticletitle{Modality-based factorization for multimodal
  fusion}.
\newblock \bibinfo{journal}{\emph{arXiv preprint arXiv:1811.12624}}
  (\bibinfo{year}{2018}).
\newblock


\bibitem[Buscaldi and Rosso(2007)]%
        {DBLP:conf/wilf/BuscaldiR07}
\bibfield{author}{\bibinfo{person}{Davide Buscaldi} {and}
  \bibinfo{person}{Paolo Rosso}.} \bibinfo{year}{2007}\natexlab{}.
\newblock \showarticletitle{Some Experiments in Humour Recognition Using the
  Italian Wikiquote Collection}. In \bibinfo{booktitle}{\emph{Applications of
  Fuzzy Sets Theory, 7th International Workshop on Fuzzy Logic and
  Applications, {WILF} 2007, Camogli, Italy, July 7-10, 2007, Proceedings}}
  \emph{(\bibinfo{series}{Lecture Notes in Computer Science},
  Vol.~\bibinfo{volume}{4578})}, \bibfield{editor}{\bibinfo{person}{Francesco
  Masulli}, \bibinfo{person}{Sushmita Mitra}, {and} \bibinfo{person}{Gabriella
  Pasi}} (Eds.). \bibinfo{publisher}{Springer}, \bibinfo{pages}{464--468}.
\newblock
\urldef\tempurl%
\url{https://doi.org/10.1007/978-3-540-73400-0\_58}
\showDOI{\tempurl}


\bibitem[Cao et~al\mbox{.}(2018)]%
        {cao2018vggface2}
\bibfield{author}{\bibinfo{person}{Qiong Cao}, \bibinfo{person}{Li Shen},
  \bibinfo{person}{Weidi Xie}, \bibinfo{person}{Omkar~M Parkhi}, {and}
  \bibinfo{person}{Andrew Zisserman}.} \bibinfo{year}{2018}\natexlab{}.
\newblock \showarticletitle{Vggface2: A dataset for recognising faces across
  pose and age}. In \bibinfo{booktitle}{\emph{2018 13th IEEE international
  conference on automatic face \& gesture recognition (FG 2018)}}. IEEE,
  \bibinfo{pages}{67--74}.
\newblock


\bibitem[Chen et~al\mbox{.}(2019)]%
        {hm11}
\bibfield{author}{\bibinfo{person}{Haifeng Chen}, \bibinfo{person}{Yifan Deng},
  \bibinfo{person}{Shiwen Cheng}, \bibinfo{person}{Yixuan Wang},
  \bibinfo{person}{Dongmei Jiang}, {and} \bibinfo{person}{Hichem Sahli}.}
  \bibinfo{year}{2019}\natexlab{}.
\newblock \showarticletitle{Efficient Spatial Temporal Convolutional Features
  for Audiovisual Continuous Affect Recognition}. In
  \bibinfo{booktitle}{\emph{Proceedings of the 9th International on
  Audio/Visual Emotion Challenge and Workshop}} (Nice, France)
  \emph{(\bibinfo{series}{AVEC '19})}. \bibinfo{publisher}{Association for
  Computing Machinery}, \bibinfo{address}{New York, NY, USA},
  \bibinfo{pages}{19–26}.
\newblock
\showISBNx{9781450369138}
\urldef\tempurl%
\url{https://doi.org/10.1145/3347320.3357690}
\showDOI{\tempurl}


\bibitem[Christ et~al\mbox{.}(2022)]%
        {baseline}
\bibfield{author}{\bibinfo{person}{Lukas Christ}, \bibinfo{person}{Shahin
  Amiriparian}, \bibinfo{person}{Alice Baird}, \bibinfo{person}{Panagiotis
  Tzirakis}, \bibinfo{person}{Alexander Kathan}, \bibinfo{person}{Niklas
  Müller}, \bibinfo{person}{Lukas Stappen}, \bibinfo{person}{Eva-Maria
  Meßner}, \bibinfo{person}{Andreas König}, \bibinfo{person}{Alan Cowen},
  \bibinfo{person}{Erik Cambria}, {and} \bibinfo{person}{Bj\"orn~W. Schuller}.}
  \bibinfo{year}{2022}\natexlab{}.
\newblock \showarticletitle{The MuSe 2022 Multimodal Sentiment Analysis
  Challenge: Humor, Emotional Reactions, and Stress}. In
  \bibinfo{booktitle}{\emph{Proceedings of the 3rd Multimodal Sentiment
  Analysis Challenge}}. \bibinfo{publisher}{Association for Computing
  Machinery}, \bibinfo{address}{Lisbon, Portugal}.
\newblock
\newblock
\shownote{Workshop held at ACM Multimedia 2022, to appear}.


\bibitem[De~Oliveira and Rodrigo(2015)]%
        {de2015humor}
\bibfield{author}{\bibinfo{person}{Luke De~Oliveira} {and}
  \bibinfo{person}{Alfredo~L Rodrigo}.} \bibinfo{year}{2015}\natexlab{}.
\newblock \showarticletitle{Humor detection in yelp reviews}.
\newblock \bibinfo{journal}{\emph{Retrieved on December}}  \bibinfo{volume}{15}
  (\bibinfo{year}{2015}), \bibinfo{pages}{2019}.
\newblock


\bibitem[Devlin et~al\mbox{.}(2018)]%
        {devlin2018bert}
\bibfield{author}{\bibinfo{person}{Jacob Devlin}, \bibinfo{person}{Ming-Wei
  Chang}, \bibinfo{person}{Kenton Lee}, {and} \bibinfo{person}{Kristina
  Toutanova}.} \bibinfo{year}{2018}\natexlab{}.
\newblock \showarticletitle{Bert: Pre-training of deep bidirectional
  transformers for language understanding}.
\newblock \bibinfo{journal}{\emph{arXiv preprint arXiv:1810.04805}}
  (\bibinfo{year}{2018}).
\newblock


\bibitem[Eyben et~al\mbox{.}(2015)]%
        {eyben2015geneva}
\bibfield{author}{\bibinfo{person}{Florian Eyben}, \bibinfo{person}{Klaus~R
  Scherer}, \bibinfo{person}{Bj{\"o}rn~W Schuller}, \bibinfo{person}{Johan
  Sundberg}, \bibinfo{person}{Elisabeth Andr{\'e}}, \bibinfo{person}{Carlos
  Busso}, \bibinfo{person}{Laurence~Y Devillers}, \bibinfo{person}{Julien
  Epps}, \bibinfo{person}{Petri Laukka}, \bibinfo{person}{Shrikanth~S
  Narayanan}, {et~al\mbox{.}}} \bibinfo{year}{2015}\natexlab{}.
\newblock \showarticletitle{The Geneva minimalistic acoustic parameter set
  (GeMAPS) for voice research and affective computing}.
\newblock \bibinfo{journal}{\emph{IEEE transactions on affective computing}}
  \bibinfo{volume}{7}, \bibinfo{number}{2} (\bibinfo{year}{2015}),
  \bibinfo{pages}{190--202}.
\newblock


\bibitem[Eyben et~al\mbox{.}(2010)]%
        {eyben2010opensmile}
\bibfield{author}{\bibinfo{person}{Florian Eyben}, \bibinfo{person}{Martin
  W{\"o}llmer}, {and} \bibinfo{person}{Bj{\"o}rn Schuller}.}
  \bibinfo{year}{2010}\natexlab{}.
\newblock \showarticletitle{Opensmile: the munich versatile and fast
  open-source audio feature extractor}. In
  \bibinfo{booktitle}{\emph{Proceedings of the 18th ACM international
  conference on Multimedia}}. \bibinfo{pages}{1459--1462}.
\newblock


\bibitem[Gu et~al\mbox{.}(2018)]%
        {related71}
\bibfield{author}{\bibinfo{person}{Yue Gu}, \bibinfo{person}{Xinyu Li},
  \bibinfo{person}{Kaixiang Huang}, \bibinfo{person}{Shiyu Fu},
  \bibinfo{person}{Kangning Yang}, \bibinfo{person}{Shuhong Chen},
  \bibinfo{person}{Moliang Zhou}, {and} \bibinfo{person}{Ivan Marsic}.}
  \bibinfo{year}{2018}\natexlab{}.
\newblock \showarticletitle{Human Conversation Analysis Using Attentive
  Multimodal Networks with Hierarchical Encoder-Decoder}. In
  \bibinfo{booktitle}{\emph{2018 {ACM} Multimedia Conference on Multimedia
  Conference, {MM} 2018, Seoul, Republic of Korea, October 22-26, 2018}},
  \bibfield{editor}{\bibinfo{person}{Susanne Boll}, \bibinfo{person}{Kyoung~Mu
  Lee}, \bibinfo{person}{Jiebo Luo}, \bibinfo{person}{Wenwu Zhu},
  \bibinfo{person}{Hyeran Byun}, \bibinfo{person}{Chang~Wen Chen},
  \bibinfo{person}{Rainer Lienhart}, {and} \bibinfo{person}{Tao Mei}} (Eds.).
  \bibinfo{publisher}{{ACM}}, \bibinfo{pages}{537--545}.
\newblock
\urldef\tempurl%
\url{https://doi.org/10.1145/3240508.3240714}
\showDOI{\tempurl}


\bibitem[Hasan et~al\mbox{.}(2019)]%
        {DBLP:conf/emnlp/HasanRZZTMH19}
\bibfield{author}{\bibinfo{person}{Md.~Kamrul Hasan}, \bibinfo{person}{Wasifur
  Rahman}, \bibinfo{person}{AmirAli~Bagher Zadeh}, \bibinfo{person}{Jianyuan
  Zhong}, \bibinfo{person}{Md.~Iftekhar Tanveer},
  \bibinfo{person}{Louis{-}Philippe Morency}, {and}
  \bibinfo{person}{Mohammed~(Ehsan) Hoque}.} \bibinfo{year}{2019}\natexlab{}.
\newblock \showarticletitle{{UR-FUNNY:} {A} Multimodal Language Dataset for
  Understanding Humor}. In \bibinfo{booktitle}{\emph{Proceedings of the 2019
  Conference on Empirical Methods in Natural Language Processing and the 9th
  International Joint Conference on Natural Language Processing, {EMNLP-IJCNLP}
  2019, Hong Kong, China, November 3-7, 2019}},
  \bibfield{editor}{\bibinfo{person}{Kentaro Inui}, \bibinfo{person}{Jing
  Jiang}, \bibinfo{person}{Vincent Ng}, {and} \bibinfo{person}{Xiaojun Wan}}
  (Eds.). \bibinfo{publisher}{Association for Computational Linguistics},
  \bibinfo{pages}{2046--2056}.
\newblock
\urldef\tempurl%
\url{https://doi.org/10.18653/v1/D19-1211}
\showDOI{\tempurl}


\bibitem[Hazarika et~al\mbox{.}(2018)]%
        {hazarika2018conversational}
\bibfield{author}{\bibinfo{person}{Devamanyu Hazarika},
  \bibinfo{person}{Soujanya Poria}, \bibinfo{person}{Amir Zadeh},
  \bibinfo{person}{Erik Cambria}, \bibinfo{person}{Louis-Philippe Morency},
  {and} \bibinfo{person}{Roger Zimmermann}.} \bibinfo{year}{2018}\natexlab{}.
\newblock \showarticletitle{Conversational memory network for emotion
  recognition in dyadic dialogue videos}. In
  \bibinfo{booktitle}{\emph{Proceedings of the conference. Association for
  Computational Linguistics. North American Chapter. Meeting}},
  Vol.~\bibinfo{volume}{2018}. NIH Public Access, \bibinfo{pages}{2122}.
\newblock


\bibitem[He et~al\mbox{.}(2016)]%
        {he2016deep}
\bibfield{author}{\bibinfo{person}{Kaiming He}, \bibinfo{person}{Xiangyu
  Zhang}, \bibinfo{person}{Shaoqing Ren}, {and} \bibinfo{person}{Jian Sun}.}
  \bibinfo{year}{2016}\natexlab{}.
\newblock \showarticletitle{Deep residual learning for image recognition}. In
  \bibinfo{booktitle}{\emph{Proceedings of the IEEE conference on computer
  vision and pattern recognition}}. \bibinfo{pages}{770--778}.
\newblock


\bibitem[Hochreiter and Schmidhuber(1997)]%
        {lstm}
\bibfield{author}{\bibinfo{person}{Sepp Hochreiter} {and}
  \bibinfo{person}{J{\"{u}}rgen Schmidhuber}.} \bibinfo{year}{1997}\natexlab{}.
\newblock \showarticletitle{Long Short-Term Memory}.
\newblock \bibinfo{journal}{\emph{Neural Comput.}} \bibinfo{volume}{9},
  \bibinfo{number}{8} (\bibinfo{year}{1997}), \bibinfo{pages}{1735--1780}.
\newblock
\urldef\tempurl%
\url{https://doi.org/10.1162/neco.1997.9.8.1735}
\showDOI{\tempurl}


\bibitem[Kampman et~al\mbox{.}(2018)]%
        {related66}
\bibfield{author}{\bibinfo{person}{Onno Kampman}, \bibinfo{person}{Elham~J.
  Barezi}, \bibinfo{person}{Dario Bertero}, {and} \bibinfo{person}{Pascale
  Fung}.} \bibinfo{year}{2018}\natexlab{}.
\newblock \showarticletitle{Investigating Audio, Visual, and Text Fusion
  Methods for End-to-End Automatic Personality Prediction}.
\newblock \bibinfo{journal}{\emph{CoRR}}  \bibinfo{volume}{abs/1805.00705}
  (\bibinfo{year}{2018}).
\newblock
\showeprint[arXiv]{1805.00705}
\urldef\tempurl%
\url{http://arxiv.org/abs/1805.00705}
\showURL{%
\tempurl}


\bibitem[Liang et~al\mbox{.}(2018)]%
        {liang2018multimodal}
\bibfield{author}{\bibinfo{person}{Paul~Pu Liang}, \bibinfo{person}{Ziyin Liu},
  \bibinfo{person}{Amir Zadeh}, {and} \bibinfo{person}{Louis-Philippe
  Morency}.} \bibinfo{year}{2018}\natexlab{}.
\newblock \showarticletitle{Multimodal language analysis with recurrent
  multistage fusion}.
\newblock \bibinfo{journal}{\emph{arXiv preprint arXiv:1808.03920}}
  (\bibinfo{year}{2018}).
\newblock


\bibitem[Liu et~al\mbox{.}(2018b)]%
        {DBLP:conf/acl/LiuZS18}
\bibfield{author}{\bibinfo{person}{Lizhen Liu}, \bibinfo{person}{Donghai
  Zhang}, {and} \bibinfo{person}{Wei Song}.} \bibinfo{year}{2018}\natexlab{b}.
\newblock \showarticletitle{Modeling Sentiment Association in Discourse for
  Humor Recognition}. In \bibinfo{booktitle}{\emph{Proceedings of the 56th
  Annual Meeting of the Association for Computational Linguistics, {ACL} 2018,
  Melbourne, Australia, July 15-20, 2018, Volume 2: Short Papers}},
  \bibfield{editor}{\bibinfo{person}{Iryna Gurevych} {and}
  \bibinfo{person}{Yusuke Miyao}} (Eds.). \bibinfo{publisher}{Association for
  Computational Linguistics}, \bibinfo{pages}{586--591}.
\newblock
\urldef\tempurl%
\url{https://doi.org/10.18653/v1/P18-2093}
\showDOI{\tempurl}


\bibitem[Liu et~al\mbox{.}(2015)]%
        {liu2015deep}
\bibfield{author}{\bibinfo{person}{Ziwei Liu}, \bibinfo{person}{Ping Luo},
  \bibinfo{person}{Xiaogang Wang}, {and} \bibinfo{person}{Xiaoou Tang}.}
  \bibinfo{year}{2015}\natexlab{}.
\newblock \showarticletitle{Deep learning face attributes in the wild}. In
  \bibinfo{booktitle}{\emph{Proceedings of the IEEE international conference on
  computer vision}}. \bibinfo{pages}{3730--3738}.
\newblock


\bibitem[Liu et~al\mbox{.}(2018a)]%
        {related20}
\bibfield{author}{\bibinfo{person}{Zhun Liu}, \bibinfo{person}{Ying Shen},
  \bibinfo{person}{Varun~Bharadhwaj Lakshminarasimhan},
  \bibinfo{person}{Paul~Pu Liang}, \bibinfo{person}{Amir Zadeh}, {and}
  \bibinfo{person}{Louis{-}Philippe Morency}.}
  \bibinfo{year}{2018}\natexlab{a}.
\newblock \showarticletitle{Efficient Low-rank Multimodal Fusion With
  Modality-Specific Factors}. In \bibinfo{booktitle}{\emph{Proceedings of the
  56th Annual Meeting of the Association for Computational Linguistics, {ACL}
  2018, Melbourne, Australia, July 15-20, 2018, Volume 1: Long Papers}},
  \bibfield{editor}{\bibinfo{person}{Iryna Gurevych} {and}
  \bibinfo{person}{Yusuke Miyao}} (Eds.). \bibinfo{publisher}{Association for
  Computational Linguistics}, \bibinfo{pages}{2247--2256}.
\newblock
\urldef\tempurl%
\url{https://doi.org/10.18653/v1/P18-1209}
\showDOI{\tempurl}


\bibitem[Ma et~al\mbox{.}(2021)]%
        {hm12}
\bibfield{author}{\bibinfo{person}{Ziyu Ma}, \bibinfo{person}{Fuyan Ma},
  \bibinfo{person}{Bin Sun}, {and} \bibinfo{person}{Shutao Li}.}
  \bibinfo{year}{2021}\natexlab{}.
\newblock \showarticletitle{Hybrid Mutimodal Fusion for Dimensional Emotion
  Recognition}. In \bibinfo{booktitle}{\emph{Proceedings of the 2nd on
  Multimodal Sentiment Analysis Challenge}} (Virtual Event, China)
  \emph{(\bibinfo{series}{MuSe '21})}. \bibinfo{publisher}{Association for
  Computing Machinery}, \bibinfo{address}{New York, NY, USA},
  \bibinfo{pages}{29–36}.
\newblock
\showISBNx{9781450386784}
\urldef\tempurl%
\url{https://doi.org/10.1145/3475957.3484457}
\showDOI{\tempurl}


\bibitem[Mao and Liu(2019)]%
        {mao2019bert}
\bibfield{author}{\bibinfo{person}{Jihang Mao} {and} \bibinfo{person}{Wanli
  Liu}.} \bibinfo{year}{2019}\natexlab{}.
\newblock \showarticletitle{A BERT-based Approach for Automatic Humor Detection
  and Scoring.}. In \bibinfo{booktitle}{\emph{IberLEF@ SEPLN}}.
  \bibinfo{pages}{197--202}.
\newblock


\bibitem[Morency et~al\mbox{.}(2011)]%
        {related15}
\bibfield{author}{\bibinfo{person}{Louis{-}Philippe Morency},
  \bibinfo{person}{Rada Mihalcea}, {and} \bibinfo{person}{Payal Doshi}.}
  \bibinfo{year}{2011}\natexlab{}.
\newblock \showarticletitle{Towards multimodal sentiment analysis: harvesting
  opinions from the web}. In \bibinfo{booktitle}{\emph{Proceedings of the 13th
  International Conference on Multimodal Interfaces, {ICMI} 2011, Alicante,
  Spain, November 14-18, 2011}}, \bibfield{editor}{\bibinfo{person}{Herv{\'{e}}
  Bourlard}, \bibinfo{person}{Thomas~S. Huang}, \bibinfo{person}{Enrique
  Vidal}, \bibinfo{person}{Daniel Gatica{-}Perez},
  \bibinfo{person}{Louis{-}Philippe Morency}, {and} \bibinfo{person}{Nicu
  Sebe}} (Eds.). \bibinfo{publisher}{{ACM}}, \bibinfo{pages}{169--176}.
\newblock
\urldef\tempurl%
\url{https://doi.org/10.1145/2070481.2070509}
\showDOI{\tempurl}


\bibitem[Nojavanasghari et~al\mbox{.}(2016)]%
        {related18}
\bibfield{author}{\bibinfo{person}{Behnaz Nojavanasghari},
  \bibinfo{person}{Deepak Gopinath}, \bibinfo{person}{Jayanth Koushik},
  \bibinfo{person}{Tadas Baltrusaitis}, {and} \bibinfo{person}{Louis{-}Philippe
  Morency}.} \bibinfo{year}{2016}\natexlab{}.
\newblock \showarticletitle{Deep multimodal fusion for persuasiveness
  prediction}. In \bibinfo{booktitle}{\emph{Proceedings of the 18th {ACM}
  International Conference on Multimodal Interaction, {ICMI} 2016, Tokyo,
  Japan, November 12-16, 2016}}, \bibfield{editor}{\bibinfo{person}{Yukiko~I.
  Nakano}, \bibinfo{person}{Elisabeth Andr{\'{e}}}, \bibinfo{person}{Toyoaki
  Nishida}, \bibinfo{person}{Louis{-}Philippe Morency}, \bibinfo{person}{Carlos
  Busso}, {and} \bibinfo{person}{Catherine Pelachaud}} (Eds.).
  \bibinfo{publisher}{{ACM}}, \bibinfo{pages}{284--288}.
\newblock
\urldef\tempurl%
\url{https://doi.org/10.1145/2993148.2993176}
\showDOI{\tempurl}


\bibitem[Poria et~al\mbox{.}(2017)]%
        {poria2017multi}
\bibfield{author}{\bibinfo{person}{Soujanya Poria}, \bibinfo{person}{Erik
  Cambria}, \bibinfo{person}{Devamanyu Hazarika}, \bibinfo{person}{Navonil
  Mazumder}, \bibinfo{person}{Amir Zadeh}, {and}
  \bibinfo{person}{Louis-Philippe Morency}.} \bibinfo{year}{2017}\natexlab{}.
\newblock \showarticletitle{Multi-level multiple attentions for contextual
  multimodal sentiment analysis}. In \bibinfo{booktitle}{\emph{2017 IEEE
  International Conference on Data Mining (ICDM)}}. IEEE,
  \bibinfo{pages}{1033--1038}.
\newblock


\bibitem[Rozgic et~al\mbox{.}(2012)]%
        {related14}
\bibfield{author}{\bibinfo{person}{Viktor Rozgic},
  \bibinfo{person}{Sankaranarayanan Ananthakrishnan}, \bibinfo{person}{Shirin
  Saleem}, \bibinfo{person}{Rohit Kumar}, {and} \bibinfo{person}{Rohit
  Prasad}.} \bibinfo{year}{2012}\natexlab{}.
\newblock \showarticletitle{Ensemble of {SVM} trees for multimodal emotion
  recognition}. In \bibinfo{booktitle}{\emph{Asia-Pacific Signal and
  Information Processing Association Annual Summit and Conference, {APSIPA}
  2012, Hollywood, CA, USA, December 3-6, 2012}}. \bibinfo{publisher}{{IEEE}},
  \bibinfo{pages}{1--4}.
\newblock
\urldef\tempurl%
\url{https://ieeexplore.ieee.org/document/6411794/}
\showURL{%
\tempurl}


\bibitem[Sun et~al\mbox{.}(2020)]%
        {sun2020multi}
\bibfield{author}{\bibinfo{person}{Licai Sun}, \bibinfo{person}{Zheng Lian},
  \bibinfo{person}{Jianhua Tao}, \bibinfo{person}{Bin Liu}, {and}
  \bibinfo{person}{Mingyue Niu}.} \bibinfo{year}{2020}\natexlab{}.
\newblock \showarticletitle{Multi-modal continuous dimensional emotion
  recognition using recurrent neural network and self-attention mechanism}. In
  \bibinfo{booktitle}{\emph{Proceedings of the 1st International on Multimodal
  Sentiment Analysis in Real-life Media Challenge and Workshop}}.
  \bibinfo{pages}{27--34}.
\newblock


\bibitem[Sun(2013)]%
        {related67}
\bibfield{author}{\bibinfo{person}{Shiliang Sun}.}
  \bibinfo{year}{2013}\natexlab{}.
\newblock \showarticletitle{A survey of multi-view machine learning}.
\newblock \bibinfo{journal}{\emph{Neural Comput. Appl.}} \bibinfo{volume}{23},
  \bibinfo{number}{7-8} (\bibinfo{year}{2013}), \bibinfo{pages}{2031--2038}.
\newblock
\urldef\tempurl%
\url{https://doi.org/10.1007/s00521-013-1362-6}
\showDOI{\tempurl}


\bibitem[Vaswani et~al\mbox{.}(2017)]%
        {attention}
\bibfield{author}{\bibinfo{person}{Ashish Vaswani}, \bibinfo{person}{Noam
  Shazeer}, \bibinfo{person}{Niki Parmar}, \bibinfo{person}{Jakob Uszkoreit},
  \bibinfo{person}{Llion Jones}, \bibinfo{person}{Aidan~N. Gomez},
  \bibinfo{person}{Lukasz Kaiser}, {and} \bibinfo{person}{Illia Polosukhin}.}
  \bibinfo{year}{2017}\natexlab{}.
\newblock \showarticletitle{Attention is All you Need}. In
  \bibinfo{booktitle}{\emph{Advances in Neural Information Processing Systems
  30: Annual Conference on Neural Information Processing Systems 2017, December
  4-9, 2017, Long Beach, CA, {USA}}},
  \bibfield{editor}{\bibinfo{person}{Isabelle Guyon}, \bibinfo{person}{Ulrike
  von Luxburg}, \bibinfo{person}{Samy Bengio}, \bibinfo{person}{Hanna~M.
  Wallach}, \bibinfo{person}{Rob Fergus}, \bibinfo{person}{S.~V.~N.
  Vishwanathan}, {and} \bibinfo{person}{Roman Garnett}} (Eds.).
  \bibinfo{pages}{5998--6008}.
\newblock
\urldef\tempurl%
\url{https://proceedings.neurips.cc/paper/2017/hash/3f5ee243547dee91fbd053c1c4a845aa-Abstract.html}
\showURL{%
\tempurl}


\bibitem[Wang et~al\mbox{.}(2017)]%
        {related65}
\bibfield{author}{\bibinfo{person}{Haohan Wang}, \bibinfo{person}{Aaksha
  Meghawat}, \bibinfo{person}{Louis-Philippe Morency}, {and}
  \bibinfo{person}{Eric~P. Xing}.} \bibinfo{year}{2017}\natexlab{}.
\newblock \showarticletitle{Select-additive learning: Improving generalization
  in multimodal sentiment analysis}. In \bibinfo{booktitle}{\emph{2017 IEEE
  International Conference on Multimedia and Expo (ICME)}}.
  \bibinfo{pages}{949--954}.
\newblock
\urldef\tempurl%
\url{https://doi.org/10.1109/ICME.2017.8019301}
\showDOI{\tempurl}


\bibitem[Wang et~al\mbox{.}(2019)]%
        {related69}
\bibfield{author}{\bibinfo{person}{Yansen Wang}, \bibinfo{person}{Ying Shen},
  \bibinfo{person}{Zhun Liu}, \bibinfo{person}{Paul~Pu Liang},
  \bibinfo{person}{Amir Zadeh}, {and} \bibinfo{person}{Louis{-}Philippe
  Morency}.} \bibinfo{year}{2019}\natexlab{}.
\newblock \showarticletitle{Words Can Shift: Dynamically Adjusting Word
  Representations Using Nonverbal Behaviors}. In \bibinfo{booktitle}{\emph{The
  Thirty-Third {AAAI} Conference on Artificial Intelligence, {AAAI} 2019, The
  Thirty-First Innovative Applications of Artificial Intelligence Conference,
  {IAAI} 2019, The Ninth {AAAI} Symposium on Educational Advances in Artificial
  Intelligence, {EAAI} 2019, Honolulu, Hawaii, USA, January 27 - February 1,
  2019}}. \bibinfo{publisher}{{AAAI} Press}, \bibinfo{pages}{7216--7223}.
\newblock
\urldef\tempurl%
\url{https://doi.org/10.1609/aaai.v33i01.33017216}
\showDOI{\tempurl}


\bibitem[W{\"{o}}llmer et~al\mbox{.}(2013)]%
        {related13}
\bibfield{author}{\bibinfo{person}{Martin W{\"{o}}llmer},
  \bibinfo{person}{Felix Weninger}, \bibinfo{person}{Tobias Knaup},
  \bibinfo{person}{Bj{\"{o}}rn~W. Schuller}, \bibinfo{person}{Congkai Sun},
  \bibinfo{person}{Kenji Sagae}, {and} \bibinfo{person}{Louis{-}Philippe
  Morency}.} \bibinfo{year}{2013}\natexlab{}.
\newblock \showarticletitle{YouTube Movie Reviews: Sentiment Analysis in an
  Audio-Visual Context}.
\newblock \bibinfo{journal}{\emph{{IEEE} Intell. Syst.}} \bibinfo{volume}{28},
  \bibinfo{number}{3} (\bibinfo{year}{2013}), \bibinfo{pages}{46--53}.
\newblock
\urldef\tempurl%
\url{https://doi.org/10.1109/MIS.2013.34}
\showDOI{\tempurl}


\bibitem[Wu and Liang(2011)]%
        {related63}
\bibfield{author}{\bibinfo{person}{Chung{-}Hsien Wu} {and}
  \bibinfo{person}{Wei{-}Bin Liang}.} \bibinfo{year}{2011}\natexlab{}.
\newblock \showarticletitle{Emotion Recognition of Affective Speech Based on
  Multiple Classifiers Using Acoustic-Prosodic Information and Semantic
  Labels}.
\newblock \bibinfo{journal}{\emph{{IEEE} Trans. Affect. Comput.}}
  \bibinfo{volume}{2}, \bibinfo{number}{1} (\bibinfo{year}{2011}),
  \bibinfo{pages}{10--21}.
\newblock
\urldef\tempurl%
\url{https://doi.org/10.1109/T-AFFC.2010.16}
\showDOI{\tempurl}


\bibitem[Yang et~al\mbox{.}(2015)]%
        {yang2015humor}
\bibfield{author}{\bibinfo{person}{Diyi Yang}, \bibinfo{person}{Alon Lavie},
  \bibinfo{person}{Chris Dyer}, {and} \bibinfo{person}{Eduard~H. Hovy}.}
  \bibinfo{year}{2015}\natexlab{}.
\newblock \showarticletitle{Humor Recognition and Humor Anchor Extraction}. In
  \bibinfo{booktitle}{\emph{Proceedings of the 2015 Conference on Empirical
  Methods in Natural Language Processing, {EMNLP} 2015, Lisbon, Portugal,
  September 17-21, 2015}}, \bibfield{editor}{\bibinfo{person}{Llu{\'{\i}}s
  M{\`{a}}rquez}, \bibinfo{person}{Chris Callison{-}Burch},
  \bibinfo{person}{Jian Su}, \bibinfo{person}{Daniele Pighin}, {and}
  \bibinfo{person}{Yuval Marton}} (Eds.). \bibinfo{publisher}{The Association
  for Computational Linguistics}, \bibinfo{pages}{2367--2376}.
\newblock
\urldef\tempurl%
\url{https://doi.org/10.18653/v1/d15-1284}
\showDOI{\tempurl}


\bibitem[Yang et~al\mbox{.}(2016)]%
        {yang2016wider}
\bibfield{author}{\bibinfo{person}{Shuo Yang}, \bibinfo{person}{Ping Luo},
  \bibinfo{person}{Chen-Change Loy}, {and} \bibinfo{person}{Xiaoou Tang}.}
  \bibinfo{year}{2016}\natexlab{}.
\newblock \showarticletitle{Wider face: A face detection benchmark}. In
  \bibinfo{booktitle}{\emph{Proceedings of the IEEE conference on computer
  vision and pattern recognition}}. \bibinfo{pages}{5525--5533}.
\newblock


\bibitem[Zadeh et~al\mbox{.}(2017)]%
        {related19}
\bibfield{author}{\bibinfo{person}{Amir Zadeh}, \bibinfo{person}{Minghai Chen},
  \bibinfo{person}{Soujanya Poria}, \bibinfo{person}{Erik Cambria}, {and}
  \bibinfo{person}{Louis{-}Philippe Morency}.} \bibinfo{year}{2017}\natexlab{}.
\newblock \showarticletitle{Tensor Fusion Network for Multimodal Sentiment
  Analysis}. In \bibinfo{booktitle}{\emph{Proceedings of the 2017 Conference on
  Empirical Methods in Natural Language Processing, {EMNLP} 2017, Copenhagen,
  Denmark, September 9-11, 2017}}, \bibfield{editor}{\bibinfo{person}{Martha
  Palmer}, \bibinfo{person}{Rebecca Hwa}, {and} \bibinfo{person}{Sebastian
  Riedel}} (Eds.). \bibinfo{publisher}{Association for Computational
  Linguistics}, \bibinfo{pages}{1103--1114}.
\newblock
\urldef\tempurl%
\url{https://doi.org/10.18653/v1/d17-1115}
\showDOI{\tempurl}


\bibitem[Zadeh et~al\mbox{.}(2018a)]%
        {related68}
\bibfield{author}{\bibinfo{person}{Amir Zadeh}, \bibinfo{person}{Paul~Pu
  Liang}, \bibinfo{person}{Navonil Mazumder}, \bibinfo{person}{Soujanya Poria},
  \bibinfo{person}{Erik Cambria}, {and} \bibinfo{person}{Louis{-}Philippe
  Morency}.} \bibinfo{year}{2018}\natexlab{a}.
\newblock \showarticletitle{Memory Fusion Network for Multi-view Sequential
  Learning}. In \bibinfo{booktitle}{\emph{Proceedings of the Thirty-Second
  {AAAI} Conference on Artificial Intelligence, (AAAI-18), the 30th innovative
  Applications of Artificial Intelligence (IAAI-18), and the 8th {AAAI}
  Symposium on Educational Advances in Artificial Intelligence (EAAI-18), New
  Orleans, Louisiana, USA, February 2-7, 2018}},
  \bibfield{editor}{\bibinfo{person}{Sheila~A. McIlraith} {and}
  \bibinfo{person}{Kilian~Q. Weinberger}} (Eds.). \bibinfo{publisher}{{AAAI}
  Press}, \bibinfo{pages}{5634--5641}.
\newblock
\urldef\tempurl%
\url{https://www.aaai.org/ocs/index.php/AAAI/AAAI18/paper/view/17341}
\showURL{%
\tempurl}


\bibitem[Zadeh et~al\mbox{.}(2018b)]%
        {related70}
\bibfield{author}{\bibinfo{person}{Amir Zadeh}, \bibinfo{person}{Paul~Pu
  Liang}, \bibinfo{person}{Soujanya Poria}, \bibinfo{person}{Prateek Vij},
  \bibinfo{person}{Erik Cambria}, {and} \bibinfo{person}{Louis{-}Philippe
  Morency}.} \bibinfo{year}{2018}\natexlab{b}.
\newblock \showarticletitle{Multi-attention Recurrent Network for Human
  Communication Comprehension}. In \bibinfo{booktitle}{\emph{Proceedings of the
  Thirty-Second {AAAI} Conference on Artificial Intelligence, (AAAI-18), the
  30th innovative Applications of Artificial Intelligence (IAAI-18), and the
  8th {AAAI} Symposium on Educational Advances in Artificial Intelligence
  (EAAI-18), New Orleans, Louisiana, USA, February 2-7, 2018}},
  \bibfield{editor}{\bibinfo{person}{Sheila~A. McIlraith} {and}
  \bibinfo{person}{Kilian~Q. Weinberger}} (Eds.). \bibinfo{publisher}{{AAAI}
  Press}, \bibinfo{pages}{5642--5649}.
\newblock
\urldef\tempurl%
\url{https://www.aaai.org/ocs/index.php/AAAI/AAAI18/paper/view/17390}
\showURL{%
\tempurl}


\bibitem[Zadeh et~al\mbox{.}(2016)]%
        {related64}
\bibfield{author}{\bibinfo{person}{Amir Zadeh}, \bibinfo{person}{Rowan
  Zellers}, \bibinfo{person}{Eli Pincus}, {and}
  \bibinfo{person}{Louis{-}Philippe Morency}.} \bibinfo{year}{2016}\natexlab{}.
\newblock \showarticletitle{{MOSI:} Multimodal Corpus of Sentiment Intensity
  and Subjectivity Analysis in Online Opinion Videos}.
\newblock \bibinfo{journal}{\emph{CoRR}}  \bibinfo{volume}{abs/1606.06259}
  (\bibinfo{year}{2016}).
\newblock
\showeprint[arXiv]{1606.06259}
\urldef\tempurl%
\url{http://arxiv.org/abs/1606.06259}
\showURL{%
\tempurl}


\bibitem[Zhang et~al\mbox{.}(2016)]%
        {zhang2016joint}
\bibfield{author}{\bibinfo{person}{Kaipeng Zhang}, \bibinfo{person}{Zhanpeng
  Zhang}, \bibinfo{person}{Zhifeng Li}, {and} \bibinfo{person}{Yu Qiao}.}
  \bibinfo{year}{2016}\natexlab{}.
\newblock \showarticletitle{Joint face detection and alignment using multitask
  cascaded convolutional networks}.
\newblock \bibinfo{journal}{\emph{IEEE signal processing letters}}
  \bibinfo{volume}{23}, \bibinfo{number}{10} (\bibinfo{year}{2016}),
  \bibinfo{pages}{1499--1503}.
\newblock


\bibitem[Zhu(2019)]%
        {DBLP:journals/cluster/Zhu19b}
\bibfield{author}{\bibinfo{person}{Dingju Zhu}.}
  \bibinfo{year}{2019}\natexlab{}.
\newblock \showarticletitle{Humor robot and humor generation method based on
  big data search through {IOT}}.
\newblock \bibinfo{journal}{\emph{Clust. Comput.}} \bibinfo{volume}{22},
  \bibinfo{number}{Supplement} (\bibinfo{year}{2019}),
  \bibinfo{pages}{9169--9175}.
\newblock
\urldef\tempurl%
\url{https://doi.org/10.1007/s10586-018-2097-z}
\showDOI{\tempurl}


\end{thebibliography}

\end{document}